\documentclass[letterpaper, 10 pt, conference]{ieeeconf} 

\IEEEoverridecommandlockouts                              

\usepackage{times}
\usepackage[pdftex]{graphicx}
\usepackage{subfigure}
\usepackage{amsmath,amsopn,amstext,amsfonts}
\usepackage{cancel}
\usepackage[space]{cite}
\usepackage{pdfsync}
\usepackage{balance}
\usepackage{color}
\usepackage{mathtools}
\usepackage{makecell}
\usepackage[linesnumbered,ruled,vlined]{algorithm2e}
\usepackage{bm}

\usepackage{diagbox}
\usepackage{float}
\usepackage{epstopdf}
\usepackage{pifont}
\usepackage{multirow}
\usepackage{url}
\usepackage{tabularx}
\usepackage[linkcolor=black,citecolor=black,urlcolor=black,colorlinks=true]{hyperref}

\usepackage{verbatim}
\usepackage{algpseudocode,algorithm2e}
\usepackage{ulem}

\newcommand{\PAR}[1]{\vskip4pt \noindent{\bf #1~}}

\SetKwInput{KwInput}{Input}                
\SetKwInput{KwOutput}{Output}              

\title{\LARGE \bf
Perceiving Unseen 3D Objects by Poking the Objects
}

\author{
    Linghao Chen
    \quad Yunzhou Song
    \quad Hujun Bao
    \quad Xiaowei Zhou 
    \\
    State Key Lab of CAD\&CG, Zhejiang University \quad
    \thanks{Corresponding author: Xiaowei Zhou.}
}

\begin{document}

    \maketitle
      \thispagestyle{empty}
    \pagestyle{empty}

    \begin{abstract}
        We present a novel approach to interactive 3D object perception for robots.
Unlike previous perception algorithms that rely on known object models or a large amount of annotated training data, we propose a poking-based approach that automatically discovers and reconstructs 3D objects.
The poking process not only enables the robot to discover unseen 3D objects but also produces multi-view observations for 3D reconstruction of the objects.
The reconstructed objects are then memorized by neural networks with regular supervised learning and can be recognized in new test images.
The experiments on real-world data show that our approach could unsupervisedly discover and reconstruct unseen 3D objects with high quality, 
and facilitate real-world applications such as robotic grasping.
The code and supplementary materials are available at the project page: \url{https://zju3dv.github.io/poking_perception/}.

    \end{abstract}

    \section{INTRODUCTION}

3D object perception plays a crucial role in computer vision and robotics, with numerous real-world applications, such as grasping, manipulation, and scene understanding.
Most existing methods for object perception either rely on known object models or a large number of annotated data for training.
Since these approaches are costly and limited to a single object instance or a few categories presented in the training data, they are hardly applicable in real-world scenarios, where many unseen objects may exist.
Imagine that a robot enters a new environment containing some objects it has never seen before, how would it perceive the 3D objects for subsequent operations?

\begin{figure}
{\centering
\resizebox{0.5\textwidth}{!}{
    \includegraphics[width=15cm,trim={7.2cm 8cm 10cm 8.5cm},clip]{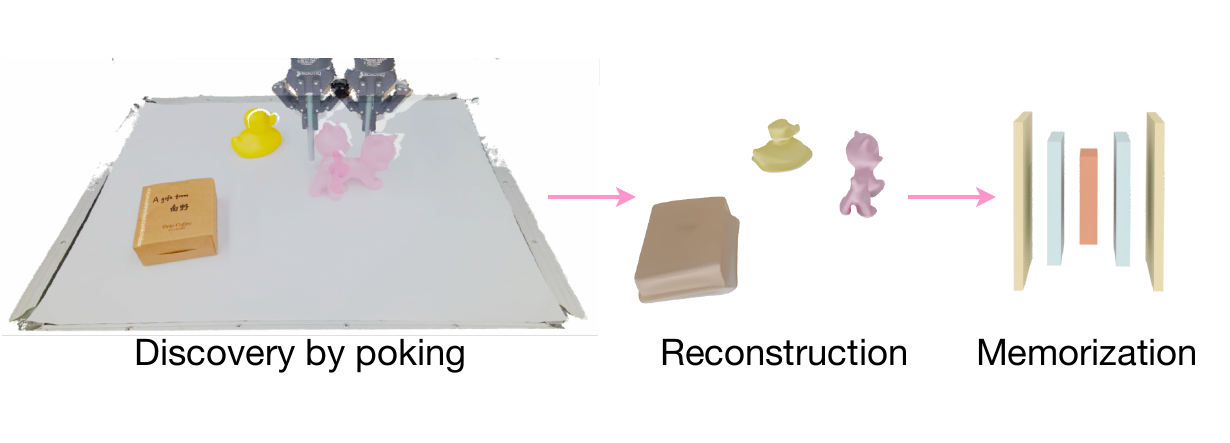}}
}
    \vspace{-2em}
    \caption{\textbf{The proposed system for 3D objects perception.}
    The poking process not only enables the system to discover unseen 3D objects but also provides multi-view observations for object reconstruction.
    Based on the reconstructed object models, the objects are memorized by neural networks for recognizing them on new test images.
    }
    \label{fig:teaser}
\end{figure}

Typically, humans understand their surroundings through interactive perception.
By interacting with objects in the scene, such as pushing, grasping, or poking, they can identify the objects and build their 3D representations, which finally serve as a knowledge base to recognize them once presented again.
In this work, we present a novel system that imitates this human behavior.
As shown in Fig.~\ref{fig:teaser}, 3D object discovery is achieved by poking, which enables the system to handle unseen 3D objects regardless of their shapes, appearances, categories, and poses.
The poking process generates multi-view observations for the 3D objects by motion, which are used to reconstruct 3D object models.
The reconstructed models are then memorized through neural networks, which are used for object recognition on new test images.

Specifically, given a scene with several unseen objects, we first generate object proposals through point cloud clustering based on geometric assumptions, which are then examined by poking with a robot arm.
The poking process prunes immovable object proposals and generates multi-view observations of 3D objects.
We then use implicit neural representation learning to reconstruct the objects based on these multi-view observations, which optimizes geometry, appearance, and poses simultaneously to yield high-quality object models.
Finally, the reconstructed models are memorized through training a detector or object pose estimator with images rendered from the models.
The memorization process allows us to recognize and perceive these objects with only one forward pass on a new test image, enabling various downstream tasks in real-world applications, such as robotic grasping, manipulation, and scene understanding.

We evaluate our system through experiments in real-world scenes.
The results show that our method can effectively discover unseen 3D objects and reconstruct them with high quality in terms of geometry, appearance, and poses.
Additionally, the memorized object models enable precise detection and pose estimation of the objects on new test images.

    \section{RELATED WORK}\label{sec:related-work}

\medskip
\noindent\textbf{Interactive perception.}
Currently, most 3D perception tasks are passive, such as object detection~\cite{ren2015faster,liu2016ssd,sun2020disp}, object pose estimation~\cite{peng2019pvnet,wang2019normalized}, object reconstruction~\cite{gkioxari2019mesh,runz2020frodo}, etc.
These methods either rely on known object models or large amounts of annotated data for training, which limits their applicability in the real world.
In contrast, several recent works in robotics propose to learn from interaction with the environment~\cite{bohg2017interactive}.
\cite{agrawal2016learning} learns to map poking to object motion by random poking and recording the change in the visual state of the world.
\cite{ye2020object, janner2018reasoning} learn the object-centric representation to build the mapping between physics actions and visual observations.
DensePhysNet~\cite{xu2019densephysnet} and DSR-Net~\cite{xu2020learning} are most relevant to ours.
DensePhysNet~\cite{xu2019densephysnet} proposes to perform a few dynamic interactions with objects to learn a dense object representation, and DSR-Net~\cite{xu2020learning} proposes to use interactive perception to discover, track, and reconstruct objects simultaneously.
However, relying on a set of pre-defined object categories or models for training limits their abilities in generalizing to unseen objects.
Recently, several works in computer vision propose to discover and perceive 3D objects by motion.
\cite{du2020unsupervised} and~\cite{kipf2021conditional} propose unsupervised training approaches to decompose the dynamic scene into the background and several moving objects using motion cues.
However, all of them struggle with real-world scenes due to the large gap between synthetic and real-world data in terms of the visual complexity and diversity of object geometries and appearances.

\medskip
\noindent\textbf{Robotic grasping.}
Traditionally, the simulator Graspit!~\cite{miller2004graspit} generates a grasp through several analytical methods given the object model.
Recent works~\cite{mahler2017dex ,morrison2018closing ,zhou2018fully ,chu2018real ,liang2019pointnetgpd ,qin2020s4g ,patten2020dgcm ,fang2020graspnet} propose learning-based approaches to learn grasping from a large amount of labeled data.
Given a depth image as input, they predict the grasp in an end-to-end manner to avoid the difficult problem of reconstructing the high-quality object model.
However, the lack of reasoning of object properties, such as geometry and semantics, limits their applicability in downstream tasks.
To tackle this problem, some methods propose to perform object reconstruction and grasping simultaneously.
\cite{van2020learning} uses the structure of the reconstruction network to classify the successful rate of grasping and use it as the objective function for continuous grasp optimization.
The reconstruction could be used to further avoid undesired contact during grasping.

\medskip
\noindent\textbf{3D reconstruction.}
Traditionally, the seminal work KinectFusion~\cite{izadi2011kinectfusion} proposes to first estimate the sensor pose using a coarse-to-fine ICP algorithm and then perform TSDF fusion~\cite{curless1996volumetric} to obtain the object geometry.
MaskFusion~\cite{runzMaskFusionRealTimeRecognition2018} and MidFusion~\cite{xuMIDFusionOctreebasedObjectLevel2018} perform instance segmentation before tracking and fusion to tackle the problem of reconstructing multiple moving objects.
Recently, implicit neural representation learning has been widely used in the 3D reconstruction.
NeRF~\cite{mildenhall2021nerf} is a pioneer work that proposes to use an MLP to predict color and density for each 3D point, which is learned by inverse volume rendering.
VolSDF~\cite{yariv2021volume} and NeuS~\cite{wang2021neus} propose to predict Signed Distance Function (SDF) instead of density to increase reconstruction quality.
\cite{yang2021objectnerf, ost2021nsg} propose to represent the scene with several neural radiance fields, each representing a foreground object or the background, to enable scene decomposition and editing.
BaRF~\cite{lin2021barf}, NeRF\verb|--|~\cite{wang2021nerfmm}, and STaR~\cite{yuan2021star} propose to jointly optimize the parameters of neural radiance fields and the relative poses between the object and the camera to reduce reliance on accurate camera/object poses in real-world applications.

    \section{METHOD}\label{sec:method}
Given a 3D scene with several objects, our goal is to enable a robot to perceive the existence and poses/geometries of the objects which are never seen before.
Our pipeline consists of three stages: we first discover the 3D objects by poking (Sec.~\ref{subsec:poking}), then reconstruct the 3D objects (Sec.~\ref{subsec:reconstruction}), and finally memorize them for recognition on new test images (Sec.~\ref{subsec:memorizing}).

\subsection{Object discovery by poking}\label{subsec:poking}
We start by describing the poking process that discovers the objects in the scene and provides input to the reconstruction module.

The poking process consists of two stages.
The first stage generates object proposals in the scene, which are then poked and examined in the second stage.

Since there exist infinite poking trajectories without any prior of object locations, we propose to first generate some object proposals and then examine them to reduce the poking search space which is analogous to the Region Proposal Network (RPN) in object detection~\cite{girshick2015fast,liu2016ssd}.
Specifically, assuming that objects are always lying on a plane, we first perform plane segmentation and then cluster the point clouds above the plane to obtain the object proposals.
The object proposals are then examined by poking and the ones which cannot be moved will be treated as negative proposals and pruned.

After generating object proposals, a robot arm pokes each object and the process is recorded using an RGB-D camera.
The design of poking trajectory only needs to ensure the objects to be viewed from an adequate number of viewpoints and avoids occlusions from the robot arm, which is achieved by performing multiple iterations of poking in a clockwise direction.
The details of the heuristic-based poking policy are described in Algorithm 1 of the supplementary material.

\noindent\textbf{Discussion.}
The utilization of learning-based grasp detection, where a neural network is employed for grasp detection followed by object grasping and scanning, is an intuitive alternative for object discovery in robotics.
However, this approach is plagued by several limitations:
1) Learning-based grasp detection is limited to the training domain and may fail on unseen objects and even damage the fragile objects;
2) Some objects may be too large to be grasped;
3) Grasping may occlude the object and make the complete reconstruction difficult.
In contrast, poking is neither limited by object categories or sizes nor does it introduce severe occlusion.
Another alternative is to obtain multi-view observations by moving a camera instead of moving the objects in the scene.
However, this approach has difficulty in segmenting objects from the scenes with complex backgrounds or when the objects are close to each other. Moreover, it cannot eliminate the occlusion between objects.
In contrast, our method effectively reduces occlusion, prunes the negative object proposals and ensures the correct number of objects thanks to the poking process.

\subsection{Object reconstruction}\label{subsec:reconstruction}

\subsubsection{Decomposed neural radiance fields}
Given the RGB-D video recorded in Sec.~\ref{subsec:poking}, we devise an implicit neural representation-based approach to reconstruct the objects.

NeRF~\cite{mildenhall2021nerf} represents a scene with a neural radiance field.
Taking as input a 3D point $\mathbf{x}$ and a viewing direction $\mathbf{d}$, a multilayer perceptron (MLP) is used to produce the density $\sigma$ and color $c$ of the point $\mathbf{x}$.
Then the pixel color along a ray is computed using volume rendering:

\begin{equation}
    \label{eq:nerf}
    \hat{C}(\mathbf{r})=\sum_{i=1}^N T_i \alpha_i \mathbf{c}_i,
\end{equation}
where $N$ is the number of 3D points along the ray $\mathbf{r}$,
$\mathbf{r}(t)= \mathbf{o} +t \mathbf{d}$ is a ray with origin $\mathbf{o}$ and direction $\mathbf{d}$, $\alpha_i = 1-\exp (-\sigma_i \delta_i)$, $T_i=\exp (-\sum_{j=1}^{i-1} \sigma_j \delta_j )$ is the accumulated transmittance along the ray, and $\delta_i=t_{i+1}-t_i$ is the distance between neighboring samples along the ray.

As a single neural radiance field could only represent one static scene,
we propose to represent our dynamic scene with a decomposed neural radiance field, in which each sub-field represents a rigid part in the scene (the background or an object) similar to ~\cite{ost2021nsg,yuan2021star}.

Meanwhile, since there is no surface constraint in the NeRF representation, we follow VolSDF~\cite{yariv2021volume} to represent the object neural radiance field as SDF and color for high-quality reconstruction.

Denoting $F^b_\Theta$ as the background NeRF, $F^k_\Theta$ as the $k$-th object VolSDF ($k=1,\cdots,K$), and $\xi^k_t \in \mathfrak{s e}(3)$ as the pose of the $k$-th object at frame $t$, for a point $\mathbf{x}$ with viewing direction $\mathbf{d}$, the color and density are computed as follows:

\begin{equation}
    \mathbf{c(x)}^b,\sigma\mathbf{(x)}^b=F^b_{\Theta}(\mathbf{x},\mathbf{d}),
\end{equation}
\begin{equation}
    \mathbf{c(x)}^k, d\mathbf{(x)}^k=F^k_{\Theta}(\mathbf{x_o},\mathbf{d}),
\end{equation}
\begin{equation}
    \sigma(\mathbf{x})^k= \begin{cases}
                              \frac{1}{\beta}\left(1-\frac{1}{2} \exp \left(\frac{d(\mathbf{x})^k}{\beta}\right)\right) & \text { if } d(\mathbf{x})^k<0 \\ \frac{1}{2 \beta} \exp \left(-\frac{d(\mathbf{x})^k}{\beta}\right) & \text { if } d(\mathbf{x})^k \geq 0 ,
    \end{cases}
\end{equation}
where $d\mathbf{(x)}^k$ is the signed distance of point $\mathbf{x}$, $\mathbf{x_o}=(\xi_t^{k})^{-1}\mathbf{x}$ is the transformed point from the world coordinate to the object coordinate, and $\beta$ is a learnable parameter.

Then, the pixel color $\hat{C}(\mathbf{r})$ and depth $\hat{D}(\mathbf{r})$ can be computed as follows:
\begin{equation}
    \hat{C}(\mathbf{r})=\sum_{i=1}^N T_i (\alpha^b_i\mathbf{c}^b_i + \sum_{k=1}^K  \alpha_i^k \mathbf{c}_i^k),
\end{equation}
\begin{equation}
    \hat D(\mathbf{r})=\sum_{i=1}^N T_i \alpha_i \mathbf{d}_i,
\end{equation}
where
$K$ is the number of neural radiance fields,
$\bar{\sigma_i}=\sigma_i^b + \sum_{k=1}^{K} \sigma_i^k$ is the composed density of all the neural radiance fields for point $\mathbf{x}_i$,
$\alpha_i = 1-\exp (-\bar{\sigma_i} \delta_i)$,
$\alpha_i^k=\frac{\sigma_i^k}{\bar{\sigma_i}} \alpha_i$,
$\alpha_i^b=\frac{\sigma_i^b}{\bar{\sigma_i}} \alpha_i$,
and $\mathbf{d}_i$ is the depth of the point $\mathbf{x}_i$.

\subsubsection{Optimizing neural radiance fields and object motion}
During optimization, we jointly optimize the parameters of the neural radiance fields $F^b_\Theta$ and $F^k_\Theta$ and the object poses $\xi^k_t$.

Given the rendered pixel color $\hat{C}(\mathbf{r})$ and depth $\hat{D}(\mathbf{r})$, we compute the color loss and depth loss as follows:
\begin{equation}
    \mathcal{L}_{\mathrm{c}}=\left\|\hat{C}(\mathbf{r})-C(\mathbf{r})\right\|,
\end{equation}
\begin{equation}
    \mathcal{L}_{\mathrm{d}}=\left\|\hat{D}(\mathbf{r})-D(\mathbf{r})\right\|,
\end{equation}
where $\left\| \cdot\right\|$ is the 1-norm, $C(\mathbf{r})$ and $D(\mathbf{r})$ are the ground-truth color and depth of ray $\mathbf{r}$.

Moreover, we apply the Eikonal loss~\cite{gropp2020implicit} to encourage $d$ to approximate a signed distance function as suggested in~\cite{yariv2021volume}.
\begin{equation}
    \mathcal{L}_{\mathrm{sdf}}=\mathbb{E}_{\boldsymbol{z}}(\|\nabla d(\boldsymbol{z})\|-1)^2,
\end{equation}

Since the object and the background are in contact, we find it hard to decompose them especially with textureless background due to its motion ambiguity.
Inspired by~\cite{yu2021plenoctrees}, we propose the following sparsity loss to solve this problem:
\begin{equation}
    \mathcal{L}_{\mathrm{sp}}=w_{sp} \left|1-\exp (-\sigma_i )\right|,
\end{equation}
where $w_{sp}=\exp(-\mathbf{w} \cdot \max(z_{m}-z_i,0))$ is the loss weight of the sparsity loss, $\sigma_i$ and $z_i$ are the density and depth of a point $x_i$ on a ray $\mathbf{r}$, $z_{m}=\max \limits_{t} \{D_\mathbf{r}^t\}$ is the maximum depth of the ray $\mathbf{r}$ across all the frames, and $\mathbf{w}$ is a weight decay parameter.

The sparsity loss encourages the density of the object VolSDF to be small, and $w_{sp}$ assigns different weights for points with different distances to the background surface.
Intuitively, the points on and farther than the background surface are assigned a large loss weight, while the points nearer than the background surface are assigned a small one.
This design eliminates the density of objects in unobserved and ambiguous spaces and reduces the effect of the sparsity loss on the spaces nearer than the background surface.

Combining the above terms, the total loss function is
\begin{equation}
    \mathcal{L}(\Theta_b,\Theta_o,\xi_o)=w_1 \mathcal{L}_\mathrm{c}+ w_2 \mathcal{L}_\mathrm{d} + w_3 \mathcal{L}_\mathrm{sdf} +w_4 \mathcal{L}_\mathrm{sp}.
\end{equation}

Once the object neural radiance field is optimized, the object mesh is extracted with the marching cubes~\cite{lorensen1987marching} operation, and the vertex colors are obtained by averaging the radiance at the vertex positions under all view directions in the input video.
The segmentation mask could be rendered by setting the radiance of the object VolSDF to 1 and the density of the background NeRF to 0.
This representation allows the network to optimize the segmentation mask implicitly and leads to a more accurate segmentation mask as demonstrated in Sec.~\ref{subsec:object-understanding-evaluation}.

\noindent\textbf{Sampling strategy.}
Since the region of the objects is relatively small compared to the entire image, we design a foreground sampling strategy for faster convergence.
Representing $N_r$ as the number of pixels to sample over an image, we propose to sample $N/2$ pixels within the object mask and the rest $N/2$ pixels over the entire image.

Moreover, we find it difficult to decompose the robot arm and objects since they are in contact during the poking process.
To restrict the impact of the robot, we propose not to sample pixels within the robot mask, which is obtained by rendering the robot arm model with its pose in each frame.

\noindent\textbf{Training strategy.}
To avoid local optima when jointly optimizing the object poses and the neural radiance fields, we initialize the object masks and poses and propose a stage-wise training strategy.
The object mask is initialized as the set of pixels whose optical flow norm is larger than a threshold.
The object poses are computed with scene flow within the object mask and Least-Squares estimation followed by Iterative Closest Points (ICP) for refinement.
The optimization process is divided into 3 stages as follows.
First, the background NeRF is initialized by sampling outside the robot arm mask and the object mask.
Second, the foreground object VolSDF is initialized by sampling only within the object mask and the object poses are fixed.
Finally, the neural radiance fields and the object poses are jointly optimized.

\subsection{Memorizing the 3D objects}\label{subsec:memorizing}
\begin{figure}
    \centering

    \resizebox{0.5\textwidth}{!}{
        \includegraphics[width=15cm,trim={3cm 10.5cm 10.5cm 5cm},clip]{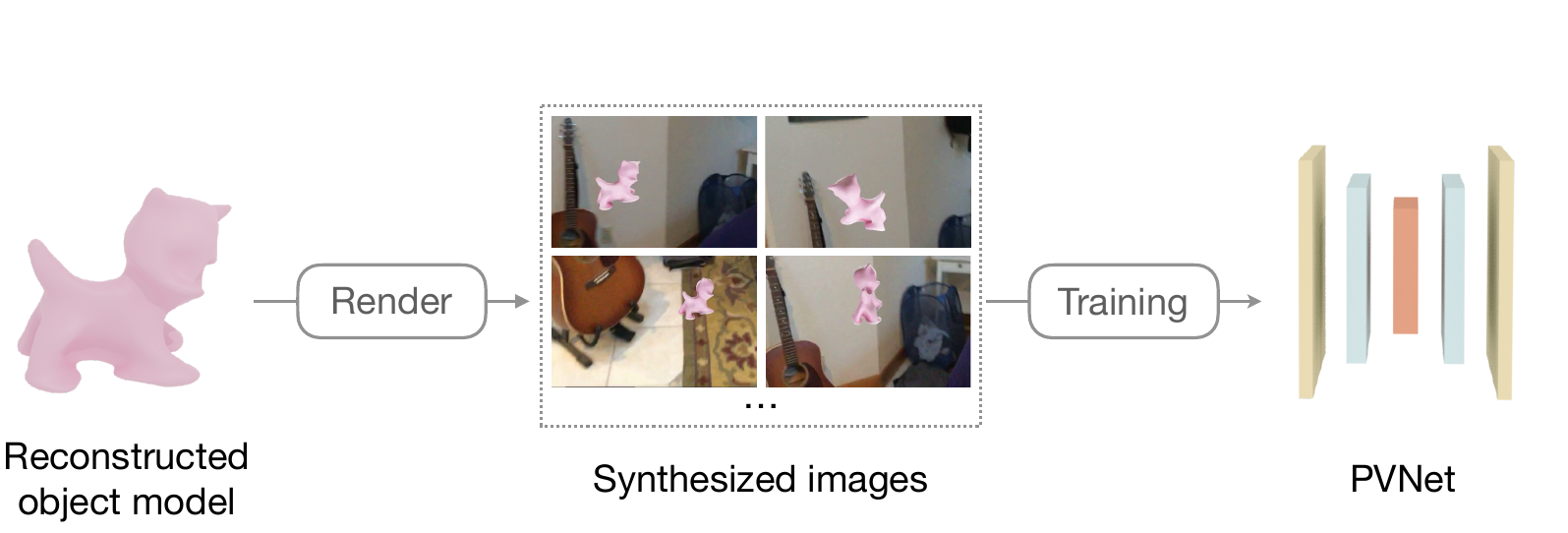}
    }
    \vspace{-2em}
    \caption{
        \textbf{The training pipeline for PVNet based on the reconstructed object model.}
        The background of the synthesized images are randomly chosen from the ScanNet dataset.
    }
    \label{fig:train_pvnet}
\end{figure}

The next step following the reconstruction is to memorize the 3D objects so that they can be rapidly recognized on new test images.
Here, we use the PVNet~\cite{peng2019pvnet} to demonstrate how to learn an object pose estimator based on the reconstructed object model.
Taking an RGB image as input, PVNet predicts the 2D keypoint positions using pixel-wise voting and computes the object pose with a Perspective-n-Point (PnP) solver~\cite{lepetit2009epnp}.
As shown in Fig.~\ref{fig:train_pvnet}, the training images for the PVNet are obtained by rendering the reconstructed model at a large number of object poses.
At inference time, ICP is used to refine the predicted object pose by aligning the reconstructed object model and the point cloud back-projected from the depth image to improve the object pose accuracy.

\subsection{Applications}\label{subsec:learning}

The perception of objects can be applied to many downstream tasks.
Here, we use robotic grasping as an example.
To grasp an object with a gripper, the relative pose between the gripper and the base of the arm is computed as follows:
\begin{figure}
    \centering

    \resizebox{0.5\textwidth}{!}{
        \includegraphics[width=15cm,trim={2.5cm 9cm 11.5cm 3.5cm},clip]{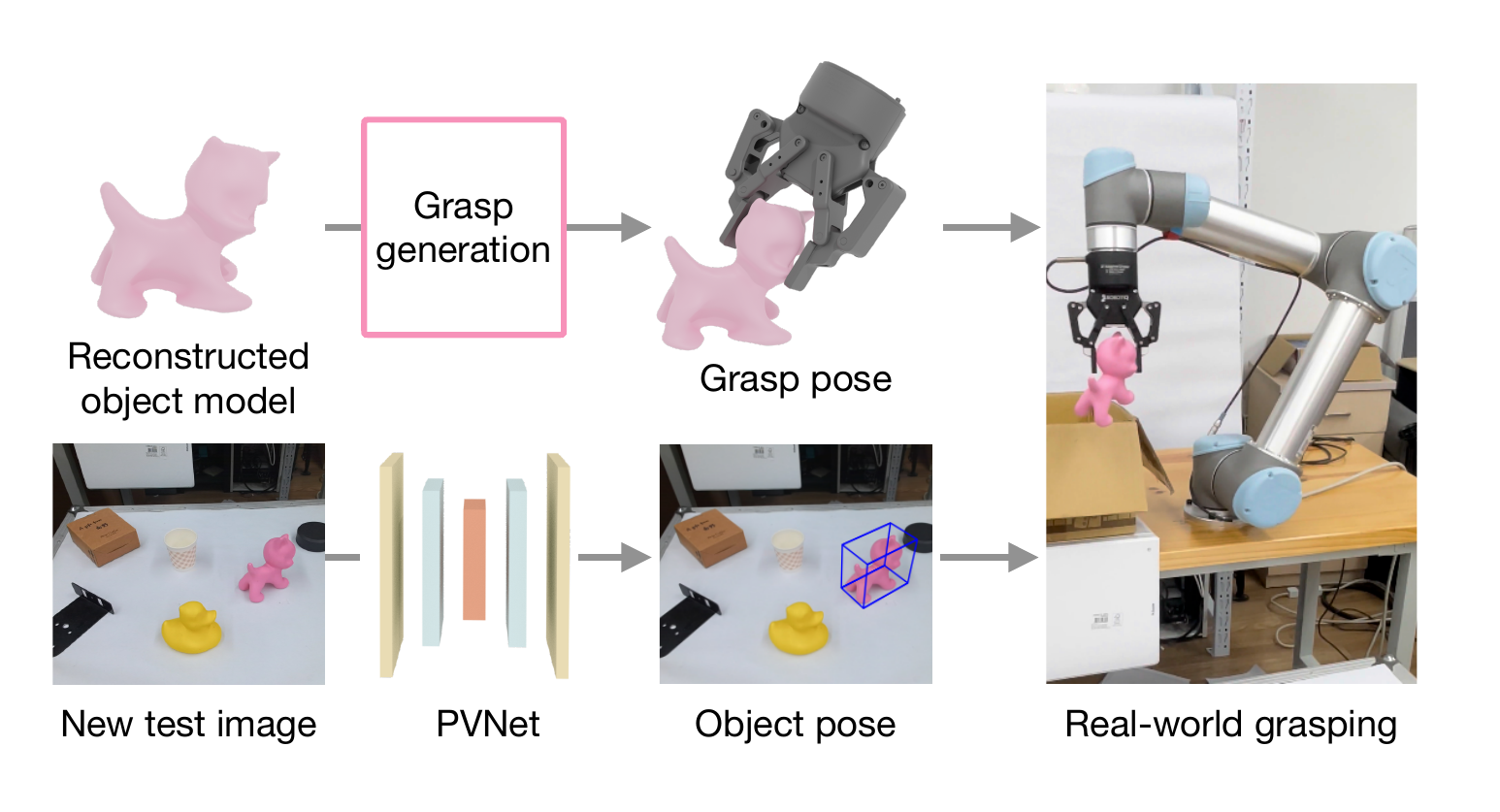}
    }
    \vspace{-2em}
    \caption{
        \textbf{Real-world grasping pipeline based on the reconstructed object model.}
        Graspit! is used to generate a grasp pose given the reconstructed object model in the object coordinates and the trained PVNet is used to estimate the object pose on the new test image.
    }
    \label{fig:real_world_grasp}
\end{figure}

\begin{equation}
    T_{gb}=T_{go} T_{oc} T_{cb},
\end{equation}
where $T_{go}$, $T_{oc}$, and $T_{cb}$ are the relative poses between the gripper and the object, the object and the camera, and the camera and the base of the arm, respectively.
As shown in Fig.~\ref{fig:real_world_grasp}, given the reconstructed object model, we use the analytic method Graspit!~\cite{miller2004graspit} to compute $T_{go}$ and PVNet~\cite{peng2019pvnet} to estimate $T_{oc}$.
$T_{cb}$ is obtained via hand-eye calibration. The details can be found in the supplementary material.

\subsection{Implementation Details}\label{subsec:implementation-details}

\medskip\noindent\textbf{Poking.}
We choose to perform 4 poking actions for each object as we empirically find this number enough to observe objects in sufficient views to obtain a complete perception.
Other details of the poking process are in the supplementary material.

\medskip\noindent\textbf{Reconstruction.}
During reconstruction, we use a batch size of 1024 rays, each sampled at 192 coordinates uniformly.
2 Adam optimizers with the learning rates decaying from 1e-3 and 5e-4 are used for the object poses and the neural radiance field parameters, respectively.
The 3 stages cost 10000, 10000, and 50000 iterations, respectively.
The loss weights are set to $w_1=1$,$w_2=1$,$w_3=0.1$, $w_4=\text{2e-5}$, and $\mathbf{w}$ is set to 200.

\medskip\noindent\textbf{Memorization.}
We synthesize 10000 images to train the PVNet.
The object poses are sampled over 30 semi-spheres with different distances to the object.
The background images are selected from the ScanNet dataset~\cite{dai2017scannet}.
To increase the generalization ability of the PVNet, both the synthesized images and the images in the recorded video are used during training.

\medskip\noindent\textbf{Grasping.}
The grasp poses are generated by the Graspit!~\cite{miller2004graspit} simulator and the one orienting downward is selected for real-world grasping to avoid collision between the gripper and the plane.
    \section{EXPERIMENTS}

\subsection{Data collection}\label{subsec:data-collection}
We capture a real-world RGB-D video to evaluate our method, where a cat, a duck, and a coffee box are put on a table.
The video consists of 665 frames.
To increase efficiency, we drop the frames with no moving objects, resulting in a 166-frame video.
The image resolution is 1344$\times$648.
The ground-truth models for the cat and the duck are provided by the LINEMOD dataset~\cite{hinterstoisser2012model}, while the coffee box is represented by a cube with manually-measured sizes.
The ground-truth object poses are obtained by aligning the object models with the RGB-D point clouds.
A mesh renderer is used to produce the ground-truth segmentation masks with the ground-truth object poses and the object models.
We recommend watching the supplementary video for the collected data.

\subsection{Object reconstruction evaluation}\label{subsec:object-understanding-evaluation}

We use MaskFusion~\cite{runz2018maskfusion} as the baseline for object reconstruction.
Since~\cite{runz2018maskfusion} cannot perform instance segmentation for unseen objects,
we use the initialized masks introduced in Sec.~\ref{subsec:reconstruction} as the masks for them.

Tab.~\ref{tab:pose_compare} compares our method with the baseline in terms of object pose accuracy.
We report the mean and maximum of rotation and translation errors.
Our method outperforms the baseline by a large margin, particularly in the maximum rotation errors for the cat and the duck, where we improved by about 20 degrees.
Our method jointly optimizes the object poses and segmentation masks for all frames, eliminating accumulated error even for textureless objects, which is not possible with ICP used in~\cite{runz2018maskfusion}.

\begin{figure}
{\centering
\resizebox{0.47\textwidth}{!}{
    \includegraphics[width=15cm,trim={2.5cm 0.2cm 3.5cm 0cm},clip]{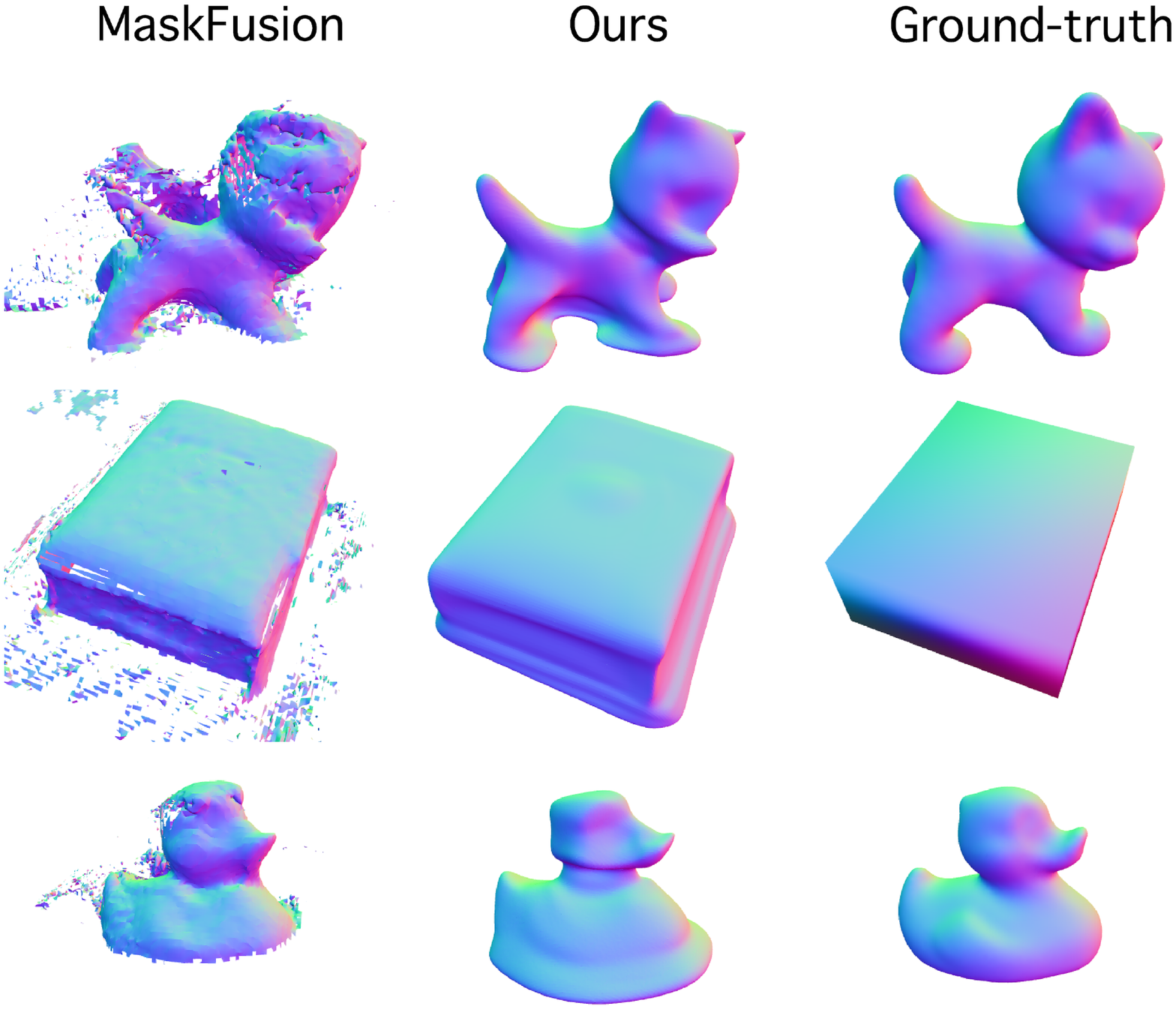}}
}
    \vspace{-1em}
    \caption{\textbf{Qualitative comparison between MaskFusion and the proposed method.}
    The color indicates surface normal.
    }
    \label{fig:mesh_compare}
\end{figure}

\begin{table}[ht]
    \centering
    \renewcommand{\arraystretch}{1.3}
    \resizebox{0.5\textwidth}{!}{
        \begin{tabular}{cccc}
            \hline
            Object                & Method & Rotation (degree)                & Translation (cm)                \\
            \hline
            \multirow{2}{*}{cat}  & MF     & 11.914 / 30.074                  & 1.676 / 4.684                   \\
            \cline{2-4}
            & Ours   & \textbf{4.391} / \textbf{8.003}  & \textbf{0.452} / \textbf{1.168} \\
            \hline
            \multirow{2}{*}{box}  & MF     & 2.060 / \textbf{3.948}           & 0.712 / \textbf{1.452}          \\
            \cline{2-4}
            & Ours   & \textbf{1.569} / 4.282           & \textbf{0.596} / 1.716          \\
            \hline
            \multirow{2}{*}{duck} & MF     & 14.144 / 31.871                  & 3.728 / 8.212                   \\
            \cline{2-4}
            & Ours   & \textbf{4.070} / \textbf{12.743} & \textbf{1.116} / \textbf{3.388} \\
            \hline\hline
        \end{tabular}
    }

    \caption{
        \textbf{Object pose comparison between MaskFusion (MF) and Ours.}
        We report mean error / maximum error over the entire video.
    }

    \label{tab:pose_compare}
\end{table}
\begin{table}[ht]
    \centering
    \renewcommand{\arraystretch}{1.3}
    \resizebox{0.5\textwidth}{!}{
        \begin{tabular}{cccccc}
            \hline
            Object                & Method & C.D. $\downarrow$ & F-score  $\uparrow$ & N.C. $\uparrow$ & Mask IoU $\uparrow$ \\
            \hline
            \multirow{2}{*}{cat}  & MF     & 0.173             & 0.836             & 0.579           & 0.708               \\
            \cline{2-6}
            & Ours   & \textbf{0.051}    & \textbf{0.926}    & \textbf{0.818}  & \textbf{0.839}      \\
            \hline
            \multirow{2}{*}{box}  & MF     & 0.705             & 0.783             & 0.657           & 0.762               \\
            \cline{2-6}
            & Ours   & \textbf{0.051}    & \textbf{0.937}    & \textbf{0.823}  & \textbf{0.790}      \\
            \hline
            \multirow{2}{*}{duck} & MF     & 0.177             & 0.812             & 0.587           & 0.674               \\
            \cline{2-6}
            & Ours   & \textbf{0.035}    & \textbf{0.963}    & \textbf{0.854}  & \textbf{0.771}      \\
            \hline\hline
        \end{tabular}
    }

    \caption{
        \textbf{3D geometry comparison between MaskFusion (MF) and Ours.}
        C.D. is chamfer distance.
        N.C. represents normal consistency.
    }
    \label{tab:mesh_compare}
\end{table}
Tab.~\ref{tab:mesh_compare} and Fig.~\ref{fig:mesh_compare} compare the results of object reconstruction and segmentation masks between our method and the baseline.
Our method outperforms the baseline in all metrics and produces higher-quality segmentation masks, especially for the cat and the duck.
This improvement is due to the joint optimization of object geometry and object pose, leading to globally consistent results.

\subsection{Object memorization evaluation}\label{subsec:object-memorization-evaluation}
\begin{figure}
{\centering
\resizebox{0.5\textwidth}{!}{
    \includegraphics[width=15cm,trim={0cm 1cm 0cm 0cm},clip]{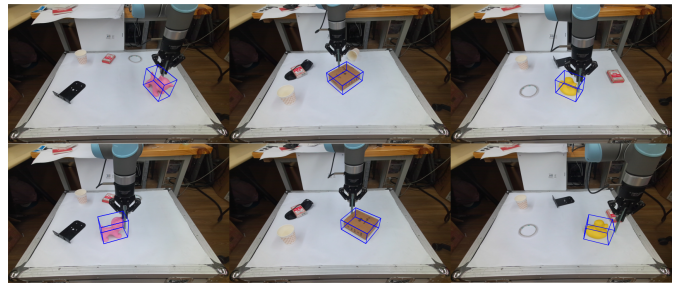}}
}
    \vspace{-2em}
    \caption{\textbf{Qualitative results of object pose estimation on new test images.}
    The estimated bounding boxes are shown in blue.
    Please refer to the supplementary video for more visualization results.
    }
    \label{fig:pp_rec_obj_pose}
\end{figure}

To evaluate object memorization, we perform object pose estimation using the trained PVNet on new test images.
Some visualization results are shown in Fig.~\ref{fig:pp_rec_obj_pose}, where the PVNet precisely estimates the object poses.

\subsection{Real-world grasping}\label{subsec:real-world-application}
\begin{figure}
{\centering
\resizebox{0.5\textwidth}{!}{
    \includegraphics[width=15cm,trim={4cm 9cm 4cm 9cm},clip]{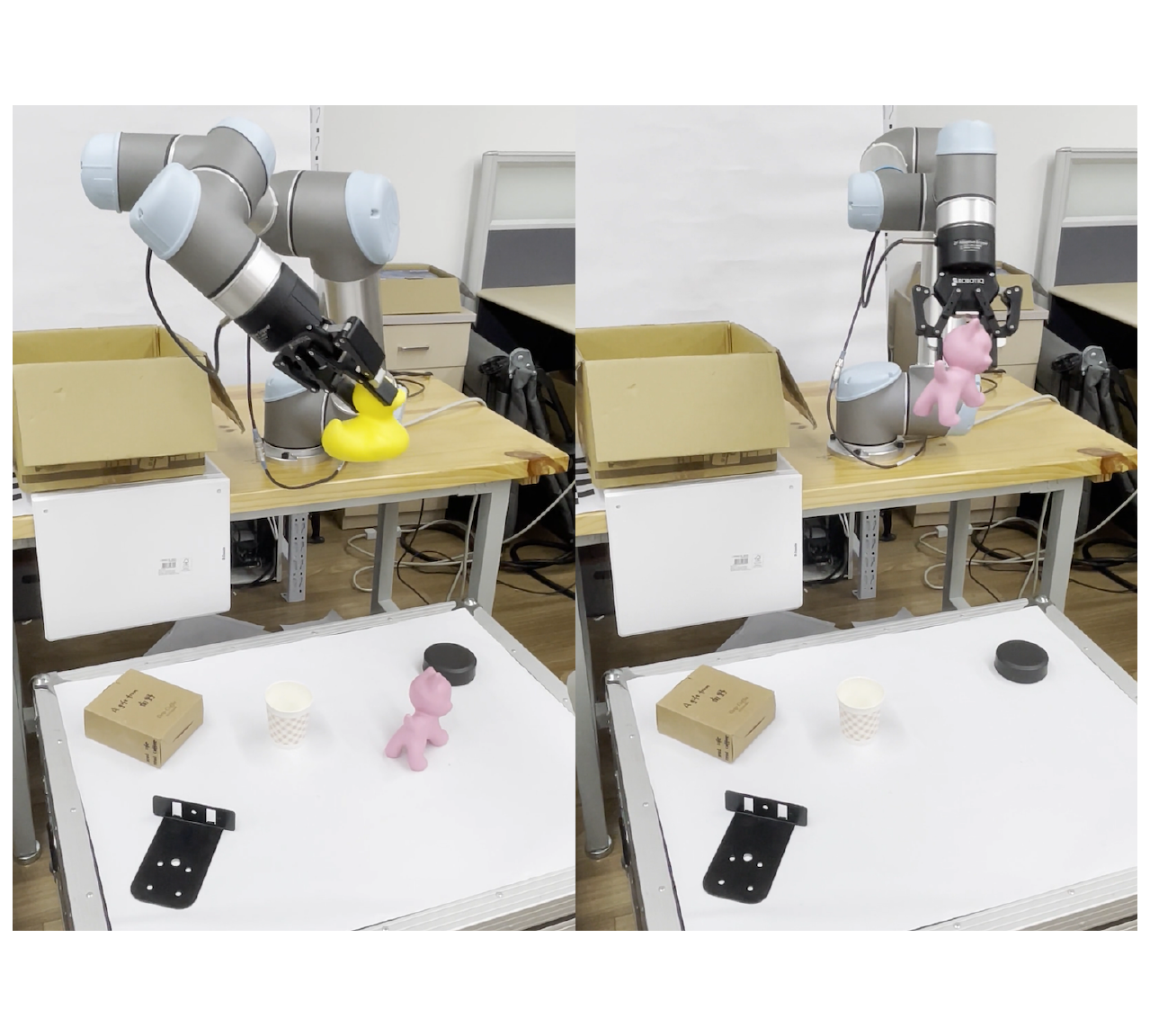}}
}
    \vspace{-2em}
    \caption{\textbf{Real-world grasping of the cat and the duck.}
    Please refer to the supplementary video for the entire grasping process.
    }
    \label{fig:grasp}
\end{figure}

We perform a real-world robotic grasping task using a parallel gripper to grasp objects placed on a plane.
The results, depicted in Fig.~\ref{fig:grasp}, show that the cat and the duck are successfully grasped.
Due to its size, the coffee box could not be grasped from the top and is not included in the demonstration.

\subsection{Ablation}\label{subsec:ablation}
\begin{figure}
    \centering
    \resizebox{0.9 \columnwidth}{!}{
        \begin{tabular}{cc}
            \includegraphics[width=0.45\linewidth,trim={4cm 0.2cm 4cm 0cm},clip]{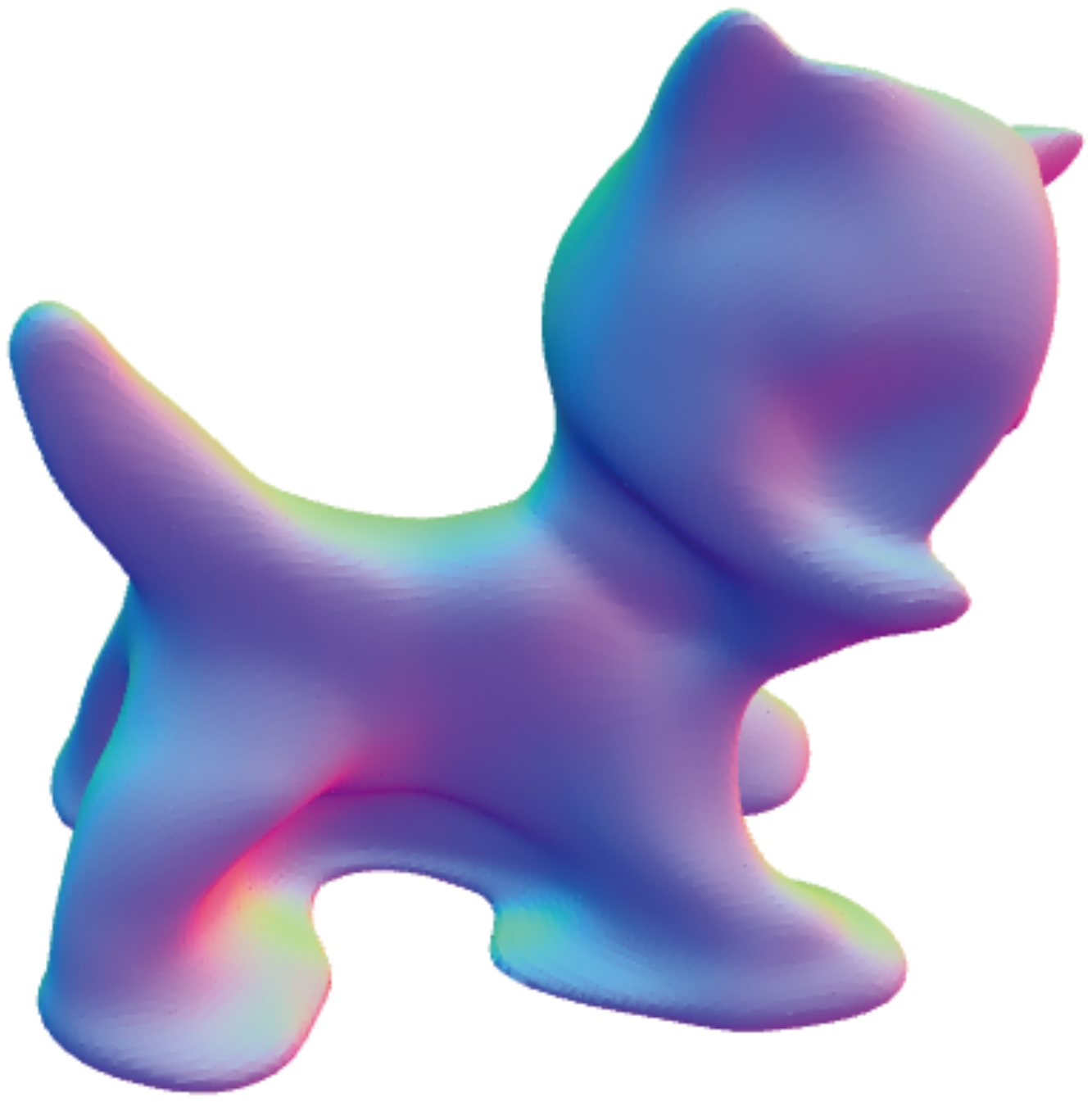} &
            \includegraphics[width=0.45\linewidth,trim={4cm 0.2cm 4cm 0cm},clip]{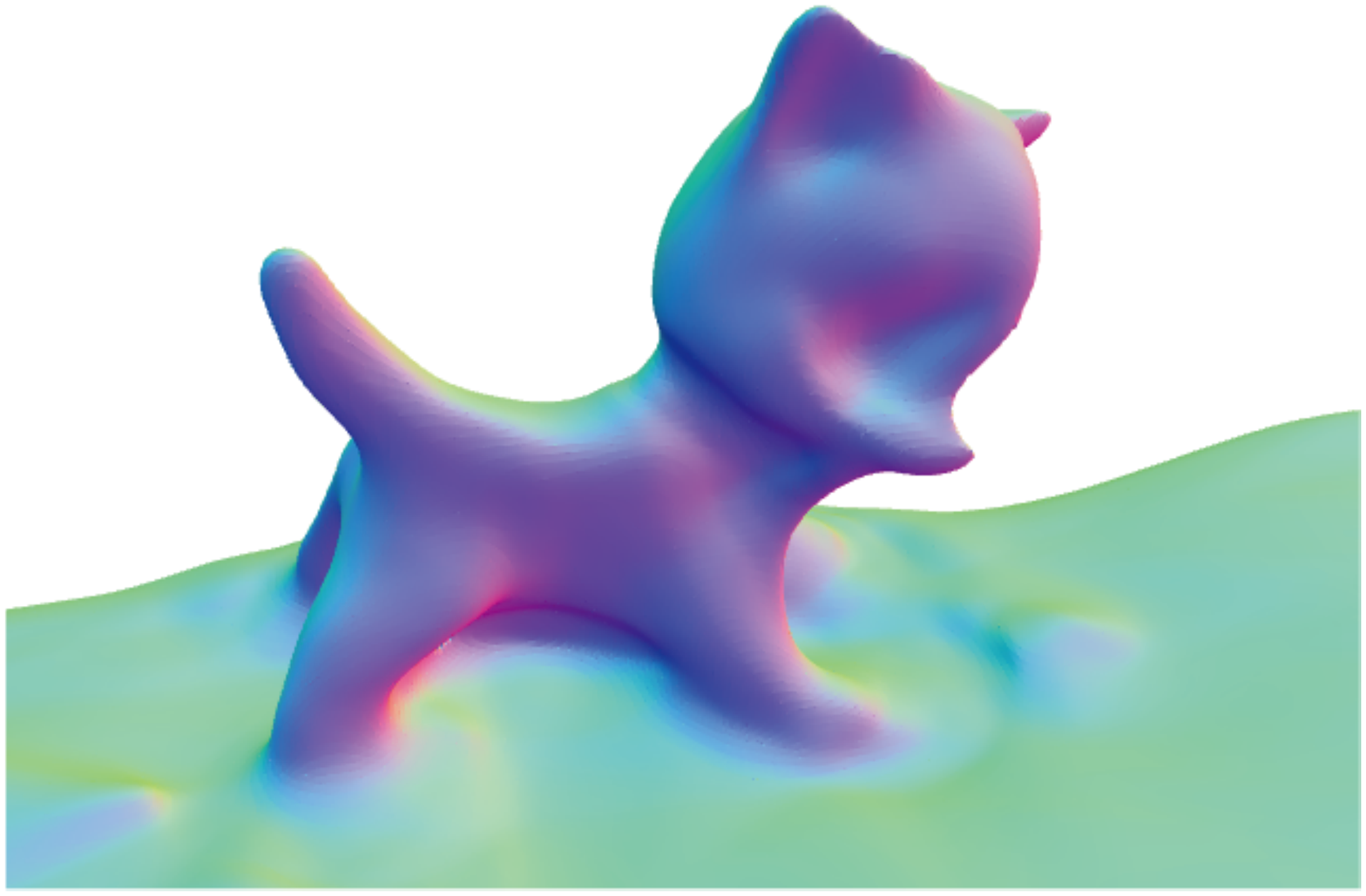}\\
            (a) & (b) \\
            \includegraphics[width=0.45\linewidth,trim={4cm 0.2cm 4cm 0cm},clip]{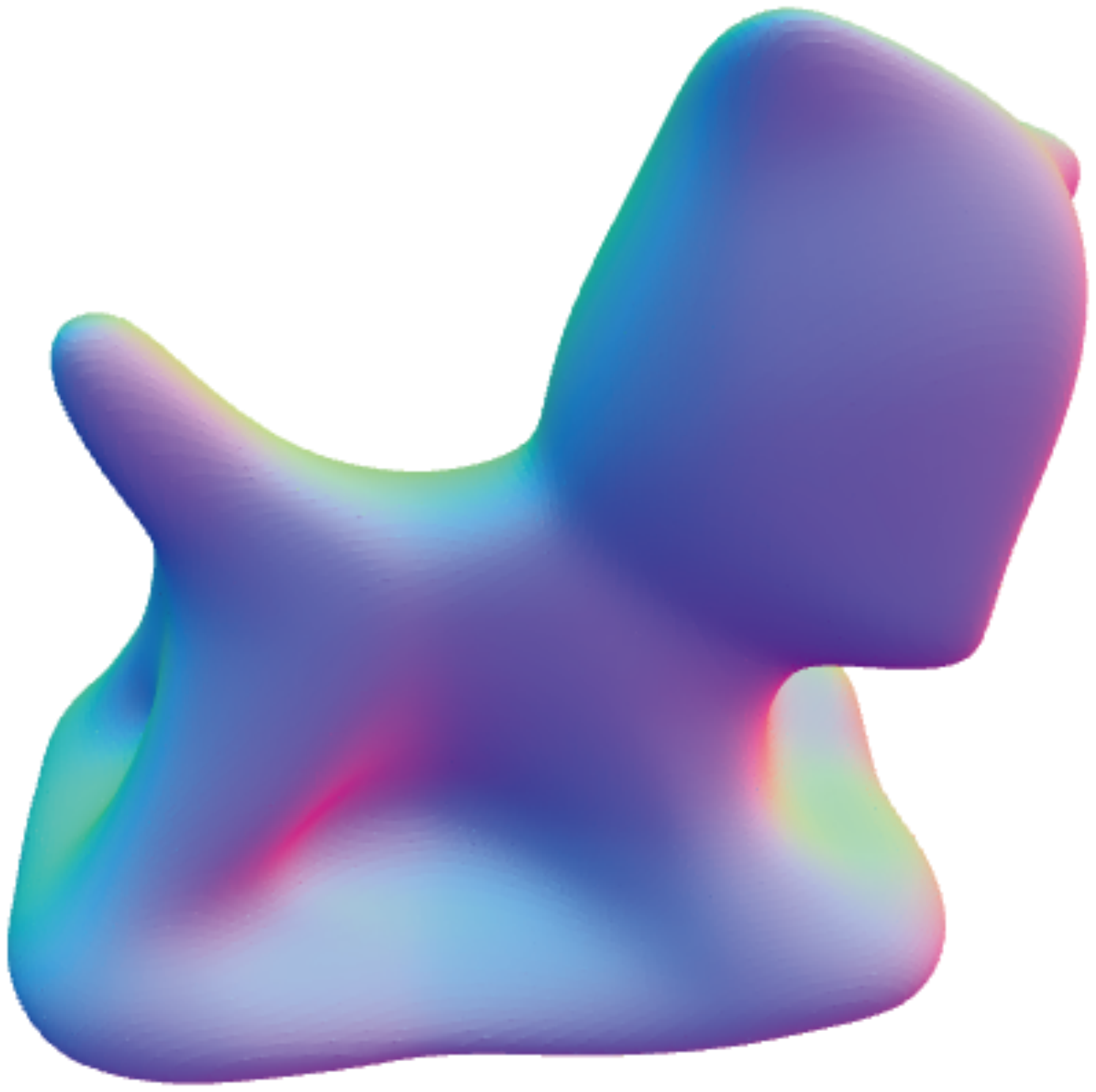} &
            \includegraphics[width=0.45\linewidth,trim={4cm 0.2cm 4cm 0cm},clip]{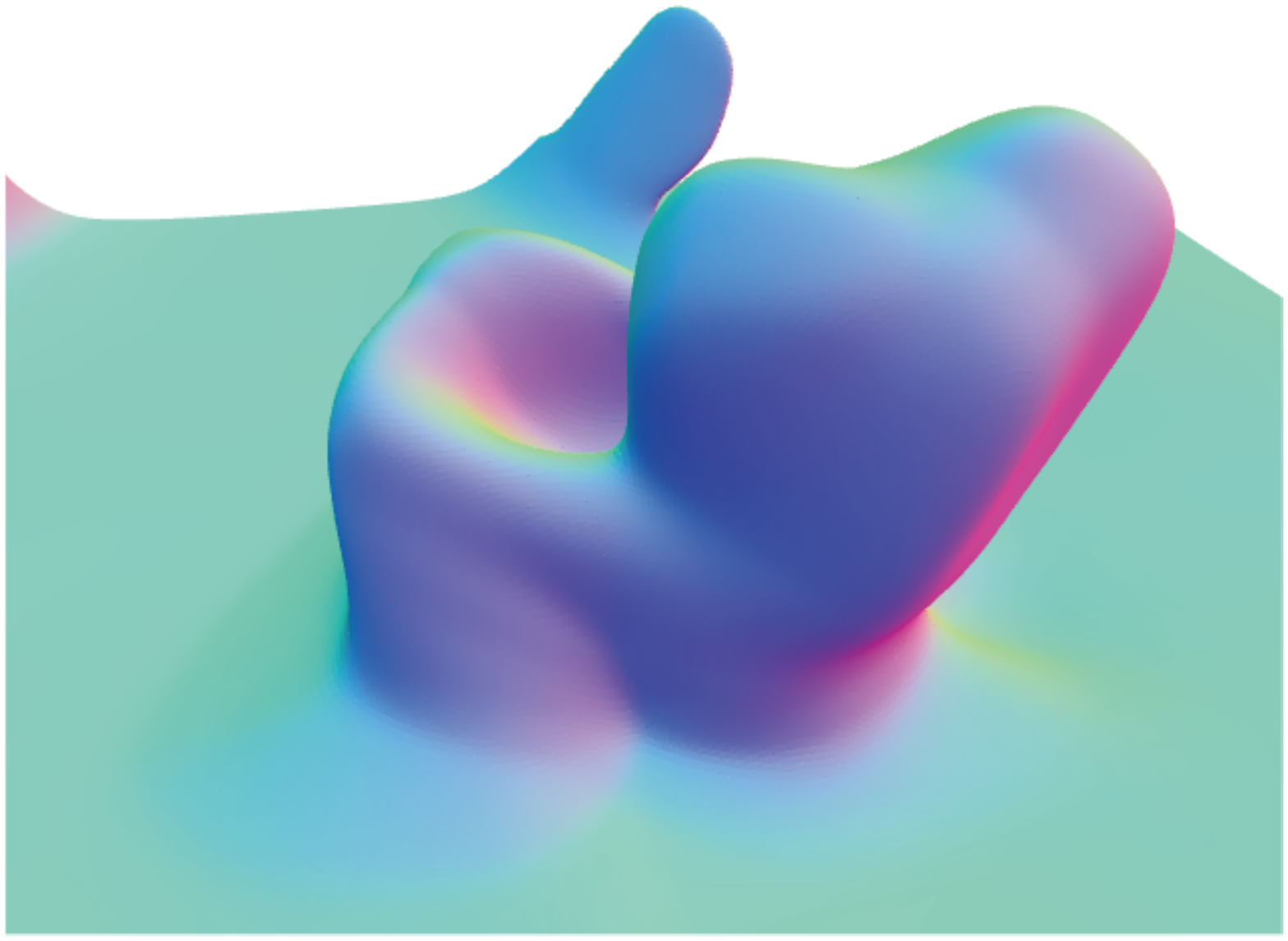}\\
            (c) & (d) \\
        \end{tabular}
    }
    \caption{\textbf{Ablation study.} Reconstructed models of
    the full version of the proposed method (a),
    without applying sparsity loss (b),
    without foreground sampling (c), 
    and without stage-wise training strategy (d) are visualized.
    }
    \label{fig:ablation}
\end{figure}

In this section, we conduct ablation experiments to analyze the effectiveness of several designs in our method.
The results of the object pose evaluation and the visualization results for the cat are shown in Tab.~\ref{tab:ablation} and Fig.~\ref{fig:ablation}, respectively.

\medskip\noindent\textbf{The sparsity loss.}
To validate the benefit of the sparsity loss, we perform optimization without sparsity loss and extract the object mesh.
As visualized in Fig.~\ref{fig:ablation} (b), our method cannot decompose the object and the background correctly without the sparsity loss due to the motion ambiguity of the texture-poor background.

\medskip\noindent\textbf{The foreground sampling strategy.}
To measure the effectiveness of the mask sampling strategy, we evaluate the performance of the proposed method with a random sampling strategy.
As shown in the second line in Tab.~\ref{tab:ablation} and Fig.~\ref{fig:ablation} (c), the optimization could not focus on the object region and thus produce very coarse results.

\medskip\noindent\textbf{The stage-wise training strategy.}
To measure the effectiveness of the stage-wise training strategy, we evaluate the performance of the proposed method with stage 3 only.
Comparing the first line and the third line in Tab.~\ref{tab:ablation} shows that the proposed method cannot decompose the foreground objects and the background correctly without initializing the radiance fields in stage 1 and stage 2.
\begin{table}[ht]
    \centering
    \renewcommand{\arraystretch}{1.3}
    \resizebox{0.5\textwidth}{!}{
        \begin{tabular}{ccc}
            \hline
            & Rotation (degree) & Translation (cm) \\
            \hline
            full               & 4.390 / 8.003     & 0.452 / 1.168    \\
            \hline
            w/o stage-wise training & 18.421 / 44.382   & 3.820 / 9.000    \\
            \hline
            w/o foreground sampling  & 9.417 / 30.385    & 1.056 / 4.160    \\
            \hline

            \hline\hline
        \end{tabular}
    }

    \caption{
        \textbf{Ablation study.}
        We report mean error / maximum error for the cat over the entire video.
    }

    \label{tab:ablation}
\end{table}

    \section{LIMITATION}\label{sec:limitation}

There are several directions to improve our system.
First, the accuracy of depth scanning is a challenge, particularly for glossy or transparent surfaces.
This leads to errors in object pose initialization and affects the accuracy of depth supervision during optimization.
Second, the current reconstruction and memorization processes are time-consuming, which can be potentially addressed with faster reconstruction methods~\cite{muller2022instant, sun2022direct, fridovich2022plenoxels} and pose estimation methods that do not require training~\cite{sun2022onepose, he2023onepose++, shugurov2022osop}.
    \section{CONCLUSIONS}

In this paper, we proposed a new system for unseen 3D object perception.
The key idea is to perform poking to discover 3D objects in the scene and then reconstruct the 3D objects based on the multi-view observations from object motion.
The reconstructed models can be then utilized to train neural networks for object recognition in new test images.
Our method achieved successful 3D object discovery and high-quality reconstruction in real-world scenarios, as demonstrated by experimental results.
The learned neural networks can be directly applied in downstream tasks like robotic grasping, manipulation, and scene understanding.
We believe that our system presents a promising approach towards the practical deployment of robots in real-world environments.

\PAR{Acknowledgement.}
The authors would like to acknowledge the support from the National Key Research and Development Program of China (No. 2020AAA0108901) and ZJU-SenseTime Joint Lab of 3D Vision.


    \addtolength{\textheight}{-1cm}   



%
%
%
%


    \normalem
    \bibliography{main}

\begin{thebibliography}{10}
\providecommand{\url}[1]{#1}
\csname url@samestyle\endcsname
\providecommand{\newblock}{\relax}
\providecommand{\bibinfo}[2]{#2}
\providecommand{\BIBentrySTDinterwordspacing}{\spaceskip=0pt\relax}
\providecommand{\BIBentryALTinterwordstretchfactor}{4}
\providecommand{\BIBentryALTinterwordspacing}{\spaceskip=\fontdimen2\font plus
\BIBentryALTinterwordstretchfactor\fontdimen3\font minus
  \fontdimen4\font\relax}
\providecommand{\BIBforeignlanguage}[2]{{%
\expandafter\ifx\csname l@#1\endcsname\relax
\typeout{** WARNING: IEEEtran.bst: No hyphenation pattern has been}%
\typeout{** loaded for the language `#1'. Using the pattern for}%
\typeout{** the default language instead.}%
\else
\language=\csname l@#1\endcsname
\fi
#2}}
\providecommand{\BIBdecl}{\relax}
\BIBdecl

\bibitem{ren2015faster}
S.~Ren, K.~He, R.~Girshick, and J.~Sun, ``Faster r-cnn: Towards real-time
  object detection with region proposal networks,'' \emph{Advances in neural
  information processing systems}, vol.~28, 2015.

\bibitem{liu2016ssd}
W.~Liu, D.~Anguelov, D.~Erhan, C.~Szegedy, S.~Reed, C.-Y. Fu, and A.~C. Berg,
  ``Ssd: Single shot multibox detector,'' in \emph{European conference on
  computer vision}.\hskip 1em plus 0.5em minus 0.4em\relax Springer, 2016, pp.
  21--37.

\bibitem{sun2020disp}
J.~Sun, L.~Chen, Y.~Xie, S.~Zhang, Q.~Jiang, X.~Zhou, and H.~Bao, ``Disp r-cnn:
  Stereo 3d object detection via shape prior guided instance disparity
  estimation,'' in \emph{Proceedings of the IEEE/CVF conference on computer
  vision and pattern recognition}, 2020, pp. 10\,548--10\,557.

\bibitem{peng2019pvnet}
S.~Peng, Y.~Liu, Q.~Huang, X.~Zhou, and H.~Bao, ``Pvnet: Pixel-wise voting
  network for 6dof pose estimation,'' in \emph{Proceedings of the IEEE/CVF
  Conference on Computer Vision and Pattern Recognition}, 2019, pp. 4561--4570.

\bibitem{wang2019normalized}
H.~Wang, S.~Sridhar, J.~Huang, J.~Valentin, S.~Song, and L.~J. Guibas,
  ``Normalized object coordinate space for category-level 6d object pose and
  size estimation,'' in \emph{Proceedings of the IEEE/CVF Conference on
  Computer Vision and Pattern Recognition}, 2019, pp. 2642--2651.

\bibitem{gkioxari2019mesh}
G.~Gkioxari, J.~Malik, and J.~Johnson, ``Mesh r-cnn,'' in \emph{Proceedings of
  the IEEE/CVF International Conference on Computer Vision}, 2019, pp.
  9785--9795.

\bibitem{runz2020frodo}
M.~Runz, K.~Li, M.~Tang, L.~Ma, C.~Kong, T.~Schmidt, I.~Reid, L.~Agapito,
  J.~Straub, S.~Lovegrove \emph{et~al.}, ``Frodo: From detections to 3d
  objects,'' in \emph{Proceedings of the IEEE/CVF Conference on Computer Vision
  and Pattern Recognition}, 2020, pp. 14\,720--14\,729.

\bibitem{bohg2017interactive}
J.~Bohg, K.~Hausman, B.~Sankaran, O.~Brock, D.~Kragic, S.~Schaal, and G.~S.
  Sukhatme, ``Interactive perception: Leveraging action in perception and
  perception in action,'' \emph{IEEE Transactions on Robotics}, vol.~33, no.~6,
  pp. 1273--1291, 2017.

\bibitem{agrawal2016learning}
P.~Agrawal, A.~V. Nair, P.~Abbeel, J.~Malik, and S.~Levine, ``Learning to poke
  by poking: Experiential learning of intuitive physics,'' \emph{Advances in
  neural information processing systems}, vol.~29, 2016.

\bibitem{ye2020object}
Y.~Ye, D.~Gandhi, A.~Gupta, and S.~Tulsiani, ``Object-centric forward modeling
  for model predictive control,'' in \emph{Conference on Robot Learning}.\hskip
  1em plus 0.5em minus 0.4em\relax PMLR, 2020, pp. 100--109.

\bibitem{janner2018reasoning}
M.~Janner, S.~Levine, W.~T. Freeman, J.~B. Tenenbaum, C.~Finn, and J.~Wu,
  ``Reasoning about physical interactions with object-oriented prediction and
  planning,'' \emph{arXiv preprint arXiv:1812.10972}, 2018.

\bibitem{xu2019densephysnet}
Z.~Xu, J.~Wu, A.~Zeng, J.~B. Tenenbaum, and S.~Song, ``Densephysnet: Learning
  dense physical object representations via multi-step dynamic interactions,''
  \emph{arXiv preprint arXiv:1906.03853}, 2019.

\bibitem{xu2020learning}
Z.~Xu, Z.~He, J.~Wu, and S.~Song, ``Learning 3d dynamic scene representations
  for robot manipulation,'' \emph{arXiv preprint arXiv:2011.01968}, 2020.

\bibitem{du2020unsupervised}
Y.~Du, K.~Smith, T.~Ulman, J.~Tenenbaum, and J.~Wu, ``Unsupervised discovery of
  3d physical objects from video,'' \emph{arXiv preprint arXiv:2007.12348},
  2020.

\bibitem{kipf2021conditional}
T.~Kipf, G.~F. Elsayed, A.~Mahendran, A.~Stone, S.~Sabour, G.~Heigold,
  R.~Jonschkowski, A.~Dosovitskiy, and K.~Greff, ``Conditional object-centric
  learning from video,'' \emph{arXiv preprint arXiv:2111.12594}, 2021.

\bibitem{miller2004graspit}
A.~T. Miller and P.~K. Allen, ``Graspit! a versatile simulator for robotic
  grasping,'' \emph{IEEE Robotics \& Automation Magazine}, vol.~11, no.~4, pp.
  110--122, 2004.

\bibitem{mahler2017dex}
J.~Mahler, J.~Liang, S.~Niyaz, M.~Laskey, R.~Doan, X.~Liu, J.~A. Ojea, and
  K.~Goldberg, ``Dex-net 2.0: Deep learning to plan robust grasps with
  synthetic point clouds and analytic grasp metrics,'' \emph{arXiv preprint
  arXiv:1703.09312}, 2017.

\bibitem{morrison2018closing}
D.~Morrison, P.~Corke, and J.~Leitner, ``Closing the loop for robotic grasping:
  A real-time, generative grasp synthesis approach,'' \emph{arXiv preprint
  arXiv:1804.05172}, 2018.

\bibitem{zhou2018fully}
X.~Zhou, X.~Lan, H.~Zhang, Z.~Tian, Y.~Zhang, and N.~Zheng, ``Fully
  convolutional grasp detection network with oriented anchor box,'' in
  \emph{2018 IEEE/RSJ International Conference on Intelligent Robots and
  Systems (IROS)}.\hskip 1em plus 0.5em minus 0.4em\relax IEEE, 2018, pp.
  7223--7230.

\bibitem{chu2018real}
F.-J. Chu, R.~Xu, and P.~A. Vela, ``Real-world multiobject, multigrasp
  detection,'' \emph{IEEE Robotics and Automation Letters}, vol.~3, no.~4, pp.
  3355--3362, 2018.

\bibitem{liang2019pointnetgpd}
H.~Liang, X.~Ma, S.~Li, M.~G{\"o}rner, S.~Tang, B.~Fang, F.~Sun, and J.~Zhang,
  ``Pointnetgpd: Detecting grasp configurations from point sets,'' in
  \emph{2019 International Conference on Robotics and Automation (ICRA)}.\hskip
  1em plus 0.5em minus 0.4em\relax IEEE, 2019, pp. 3629--3635.

\bibitem{qin2020s4g}
Y.~Qin, R.~Chen, H.~Zhu, M.~Song, J.~Xu, and H.~Su, ``S4g: Amodal single-view
  single-shot se (3) grasp detection in cluttered scenes,'' in \emph{Conference
  on robot learning}.\hskip 1em plus 0.5em minus 0.4em\relax PMLR, 2020, pp.
  53--65.

\bibitem{patten2020dgcm}
T.~Patten, K.~Park, and M.~Vincze, ``Dgcm-net: dense geometrical correspondence
  matching network for incremental experience-based robotic grasping,''
  \emph{Frontiers in Robotics and AI}, vol.~7, p. 120, 2020.

\bibitem{fang2020graspnet}
H.-S. Fang, C.~Wang, M.~Gou, and C.~Lu, ``Graspnet-1billion: A large-scale
  benchmark for general object grasping,'' in \emph{Proceedings of the IEEE/CVF
  conference on computer vision and pattern recognition}, 2020, pp.
  11\,444--11\,453.

\bibitem{van2020learning}
M.~Van~der Merwe, Q.~Lu, B.~Sundaralingam, M.~Matak, and T.~Hermans, ``Learning
  continuous 3d reconstructions for geometrically aware grasping,'' in
  \emph{2020 IEEE International Conference on Robotics and Automation
  (ICRA)}.\hskip 1em plus 0.5em minus 0.4em\relax IEEE, 2020, pp.
  11\,516--11\,522.

\bibitem{izadi2011kinectfusion}
S.~Izadi, D.~Kim, O.~Hilliges, D.~Molyneaux, R.~Newcombe, P.~Kohli, J.~Shotton,
  S.~Hodges, D.~Freeman, A.~Davison \emph{et~al.}, ``Kinectfusion: real-time 3d
  reconstruction and interaction using a moving depth camera,'' in
  \emph{Proceedings of the 24th annual ACM symposium on User interface software
  and technology}, 2011, pp. 559--568.

\bibitem{curless1996volumetric}
B.~Curless and M.~Levoy, ``A volumetric method for building complex models from
  range images,'' in \emph{Proceedings of the 23rd annual conference on
  Computer graphics and interactive techniques}, 1996, pp. 303--312.

\bibitem{runzMaskFusionRealTimeRecognition2018}
M.~Runz, M.~Buffier, and L.~Agapito, ``Maskfusion: Real-time recognition,
  tracking and reconstruction of multiple moving objects,'' in \emph{2018 IEEE
  International Symposium on Mixed and Augmented Reality (ISMAR)}.\hskip 1em
  plus 0.5em minus 0.4em\relax IEEE, 2018, pp. 10--20.

\bibitem{xuMIDFusionOctreebasedObjectLevel2018}
B.~Xu, W.~Li, D.~Tzoumanikas, M.~Bloesch, A.~Davison, and S.~Leutenegger,
  ``Mid-fusion: Octree-based object-level multi-instance dynamic slam,'' in
  \emph{2019 International Conference on Robotics and Automation (ICRA)}.\hskip
  1em plus 0.5em minus 0.4em\relax IEEE, 2019, pp. 5231--5237.

\bibitem{mildenhall2021nerf}
B.~Mildenhall, P.~P. Srinivasan, M.~Tancik, J.~T. Barron, R.~Ramamoorthi, and
  R.~Ng, ``Nerf: Representing scenes as neural radiance fields for view
  synthesis,'' \emph{Communications of the ACM}, vol.~65, no.~1, pp. 99--106,
  2021.

\bibitem{yariv2021volume}
L.~Yariv, J.~Gu, Y.~Kasten, and Y.~Lipman, ``Volume rendering of neural
  implicit surfaces,'' \emph{Advances in Neural Information Processing
  Systems}, vol.~34, pp. 4805--4815, 2021.

\bibitem{wang2021neus}
P.~Wang, L.~Liu, Y.~Liu, C.~Theobalt, T.~Komura, and W.~Wang, ``Neus: Learning
  neural implicit surfaces by volume rendering for multi-view reconstruction,''
  \emph{arXiv preprint arXiv:2106.10689}, 2021.

\bibitem{yang2021objectnerf}
B.~Yang, Y.~Zhang, Y.~Xu, Y.~Li, H.~Zhou, H.~Bao, G.~Zhang, and Z.~Cui,
  ``Learning object-compositional neural radiance field for editable scene
  rendering,'' in \emph{Proceedings of the IEEE/CVF International Conference on
  Computer Vision}, 2021, pp. 13\,779--13\,788.

\bibitem{ost2021nsg}
J.~Ost, F.~Mannan, N.~Thuerey, J.~Knodt, and F.~Heide, ``Neural scene graphs
  for dynamic scenes,'' in \emph{Proceedings of the IEEE/CVF Conference on
  Computer Vision and Pattern Recognition}, 2021, pp. 2856--2865.

\bibitem{lin2021barf}
C.-H. Lin, W.-C. Ma, A.~Torralba, and S.~Lucey, ``Barf: Bundle-adjusting neural
  radiance fields,'' in \emph{Proceedings of the IEEE/CVF International
  Conference on Computer Vision}, 2021, pp. 5741--5751.

\bibitem{wang2021nerfmm}
Z.~Wang, S.~Wu, W.~Xie, M.~Chen, and V.~A. Prisacariu, ``Nerf--: Neural
  radiance fields without known camera parameters,'' \emph{arXiv preprint
  arXiv:2102.07064}, 2021.

\bibitem{yuan2021star}
W.~Yuan, Z.~Lv, T.~Schmidt, and S.~Lovegrove, ``Star: Self-supervised tracking
  and reconstruction of rigid objects in motion with neural rendering,'' in
  \emph{Proceedings of the IEEE/CVF Conference on Computer Vision and Pattern
  Recognition}, 2021, pp. 13\,144--13\,152.

\bibitem{girshick2015fast}
R.~Girshick, ``Fast r-cnn,'' in \emph{Proceedings of the IEEE international
  conference on computer vision}, 2015, pp. 1440--1448.

\bibitem{gropp2020implicit}
A.~Gropp, L.~Yariv, N.~Haim, M.~Atzmon, and Y.~Lipman, ``Implicit geometric
  regularization for learning shapes,'' \emph{arXiv preprint arXiv:2002.10099},
  2020.

\bibitem{yu2021plenoctrees}
A.~Yu, R.~Li, M.~Tancik, H.~Li, R.~Ng, and A.~Kanazawa, ``Plenoctrees for
  real-time rendering of neural radiance fields,'' in \emph{Proceedings of the
  IEEE/CVF International Conference on Computer Vision}, 2021, pp. 5752--5761.

\bibitem{lorensen1987marching}
W.~E. Lorensen and H.~E. Cline, ``Marching cubes: A high resolution 3d surface
  construction algorithm,'' \emph{ACM siggraph computer graphics}, vol.~21,
  no.~4, pp. 163--169, 1987.

\bibitem{lepetit2009epnp}
V.~Lepetit, F.~Moreno-Noguer, and P.~Fua, ``Epnp: An accurate o (n) solution to
  the pnp problem,'' \emph{International journal of computer vision}, vol.~81,
  no.~2, pp. 155--166, 2009.

\bibitem{dai2017scannet}
A.~Dai, A.~X. Chang, M.~Savva, M.~Halber, T.~Funkhouser, and M.~Nie{\ss}ner,
  ``Scannet: Richly-annotated 3d reconstructions of indoor scenes,'' in
  \emph{Proc. Computer Vision and Pattern Recognition (CVPR), IEEE}, 2017.

\bibitem{hinterstoisser2012model}
S.~Hinterstoisser, V.~Lepetit, S.~Ilic, S.~Holzer, G.~Bradski, K.~Konolige, and
  N.~Navab, ``Model based training, detection and pose estimation of
  texture-less 3d objects in heavily cluttered scenes,'' in \emph{Asian
  conference on computer vision}.\hskip 1em plus 0.5em minus 0.4em\relax
  Springer, 2012, pp. 548--562.

\bibitem{runz2018maskfusion}
M.~Runz, M.~Buffier, and L.~Agapito, ``Maskfusion: Real-time recognition,
  tracking and reconstruction of multiple moving objects,'' in \emph{2018 IEEE
  International Symposium on Mixed and Augmented Reality (ISMAR)}.\hskip 1em
  plus 0.5em minus 0.4em\relax IEEE, 2018, pp. 10--20.

\bibitem{muller2022instant}
T.~M{\"u}ller, A.~Evans, C.~Schied, and A.~Keller, ``Instant neural graphics
  primitives with a multiresolution hash encoding,'' \emph{arXiv preprint
  arXiv:2201.05989}, 2022.

\bibitem{sun2022direct}
C.~Sun, M.~Sun, and H.-T. Chen, ``Direct voxel grid optimization: Super-fast
  convergence for radiance fields reconstruction,'' in \emph{Proceedings of the
  IEEE/CVF Conference on Computer Vision and Pattern Recognition}, 2022, pp.
  5459--5469.

\bibitem{fridovich2022plenoxels}
S.~Fridovich-Keil, A.~Yu, M.~Tancik, Q.~Chen, B.~Recht, and A.~Kanazawa,
  ``Plenoxels: Radiance fields without neural networks,'' in \emph{Proceedings
  of the IEEE/CVF Conference on Computer Vision and Pattern Recognition}, 2022,
  pp. 5501--5510.

\bibitem{sun2022onepose}
J.~Sun, Z.~Wang, S.~Zhang, X.~He, H.~Zhao, G.~Zhang, and X.~Zhou, ``Onepose:
  One-shot object pose estimation without cad models,'' in \emph{Proceedings of
  the IEEE/CVF Conference on Computer Vision and Pattern Recognition}, 2022,
  pp. 6825--6834.

\bibitem{he2023onepose++}
X.~He, J.~Sun, Y.~Wang, D.~Huang, H.~Bao, and X.~Zhou, ``Onepose++:
  Keypoint-free one-shot object pose estimation without cad models,''
  \emph{arXiv preprint arXiv:2301.07673}, 2023.

\bibitem{shugurov2022osop}
I.~Shugurov, F.~Li, B.~Busam, and S.~Ilic, ``Osop: A multi-stage one shot
  object pose estimation framework,'' in \emph{Proceedings of the IEEE/CVF
  Conference on Computer Vision and Pattern Recognition}, 2022, pp. 6835--6844.

\end{thebibliography}


\begin{thebibliography}{1}
\providecommand{\url}[1]{#1}
\csname url@samestyle\endcsname
\providecommand{\newblock}{\relax}
\providecommand{\bibinfo}[2]{#2}
\providecommand{\BIBentrySTDinterwordspacing}{\spaceskip=0pt\relax}
\providecommand{\BIBentryALTinterwordstretchfactor}{4}
\providecommand{\BIBentryALTinterwordspacing}{\spaceskip=\fontdimen2\font plus
\BIBentryALTinterwordstretchfactor\fontdimen3\font minus
  \fontdimen4\font\relax}
\providecommand{\BIBforeignlanguage}[2]{{%
\expandafter\ifx\csname l@#1\endcsname\relax
\typeout{** WARNING: IEEEtran.bst: No hyphenation pattern has been}%
\typeout{** loaded for the language `#1'. Using the pattern for}%
\typeout{** the default language instead.}%
\else
\language=\csname l@#1\endcsname
\fi
#2}}
\providecommand{\BIBdecl}{\relax}
\BIBdecl

\bibitem{teed2020raft}
Z.~Teed and J.~Deng, ``Raft: Recurrent all-pairs field transforms for optical
  flow,'' in \emph{European conference on computer vision}.\hskip 1em plus
  0.5em minus 0.4em\relax Springer, 2020, pp. 402--419.

\bibitem{kato2018neural}
H.~Kato, Y.~Ushiku, and T.~Harada, ``Neural 3d mesh renderer,'' in
  \emph{Proceedings of the IEEE conference on computer vision and pattern
  recognition}, 2018, pp. 3907--3916.

\end{thebibliography}

\end{document}


\title{Supplementary Material for\\ \LARGE \bf
    Perceiving Unseen 3D Objects by Poking the Objects}
    \maketitle

    \section{Poking policy}\label{sec:poking-policy}
    \begin{algorithm}
    \DontPrintSemicolon

    \KwInput{Object point cloud $P_o$;
    start distance $d_s$; poking distance $d_p$;
    poking iterations $N_p$
    }
    \KwOutput{start point $s$; end point $e$}
    Fit oriented bounding box $B=(b_1,b_2,b_3,b_4)$ in bird-eye-view for $P_o$\\
    \For{$n_p$ = 1 to $N_p$}{

        sort $B$ by $\left\{b_{ix}\right\}$ increasingly \\
        \If{$\left| b_1 - b_2 \right| < \left| b_1 - b_3 \right|$}
        {
            $p_1, p_2 \leftarrow b_1,b_3$
        }
        \Else
        {
            $p_1, p_2 \leftarrow b_1,b_2$
        }
        $r \leftarrow 75 ^{\circ}$ \\
        \If{$p_{1y} < p_{2y}$}
        {
            $p_1,p_2 \leftarrow p_2, p_1$\\
            $r \leftarrow 105 ^{\circ}$
        }

        $d \leftarrow
            \left[\begin{array}{cc}
                        \cos r & -\sin r \\
                        \sin r & \cos r
            \end{array}\right] \left[\begin{array}{c} p_{2x}-p_{1x}\\
                                         p_{2y}-p_{1y}
            \end{array}\right]$

        $p \leftarrow p_1+ (p_2-p_1)/3$ \\

        $s \leftarrow p-d \cdot d_s$ \\
        $e \leftarrow p+d \cdot d_p$ \\

        Poke from $s$ to $e$ \\
        Update $B$\\
    }

    \caption{Poking algorithm.}\label{alg:poking_policy2}
\end{algorithm}
    The coordinate systems are defined as follows: the $z$ axes of the world and the robot are perpendicular to the plane and the $y$ axes are perpendicular to the line connecting the camera and the center of the world coordinate system.
    The line $p_1 p_2$ represents one of the longer sides of the bounding box $B$ that rotates the object in a clockwise direction upon poking.
    The angle $r$ between the poking direction and the longer side of the bounding box is defined in Algorithm 1 (lines 8 and 11).
    Instead of poking in a perpendicular direction ($r=90 ^{\circ}$), we empirically add or subtract $15 ^{\circ}$ to achieve more translation of the object.
    The $z$ values of the start and end points are determined by keeping the distance between the end of the gripper and the plane constant.
    Before execution, the computed start and end points are evaluated to avoid any unexpected collisions between the gripper and the environment.
    After each iteration of poking, the oriented bounding box $B$ is updated through point cloud registration on the foreground, which is defined as the set of pixels with optical flow norms greater than a pre-defined threshold.
    The point cloud registration is achieved through scene flow estimation and refined by the Iterative Closest Point (ICP) method.
    The scene flow is obtained from the back-projected optical flow, which is estimated using a trained optical flow estimation network~\cite{teed2020raft}.
    The order of poking is determined by sorting the distances between the center of the object proposal and the right-down corner of the plane in increasing order.

    \section{Data collection}\label{sec:data-collection}
    \begin{figure}
{\centering
\resizebox{0.5\textwidth}{!}{
    \includegraphics[width=15cm]{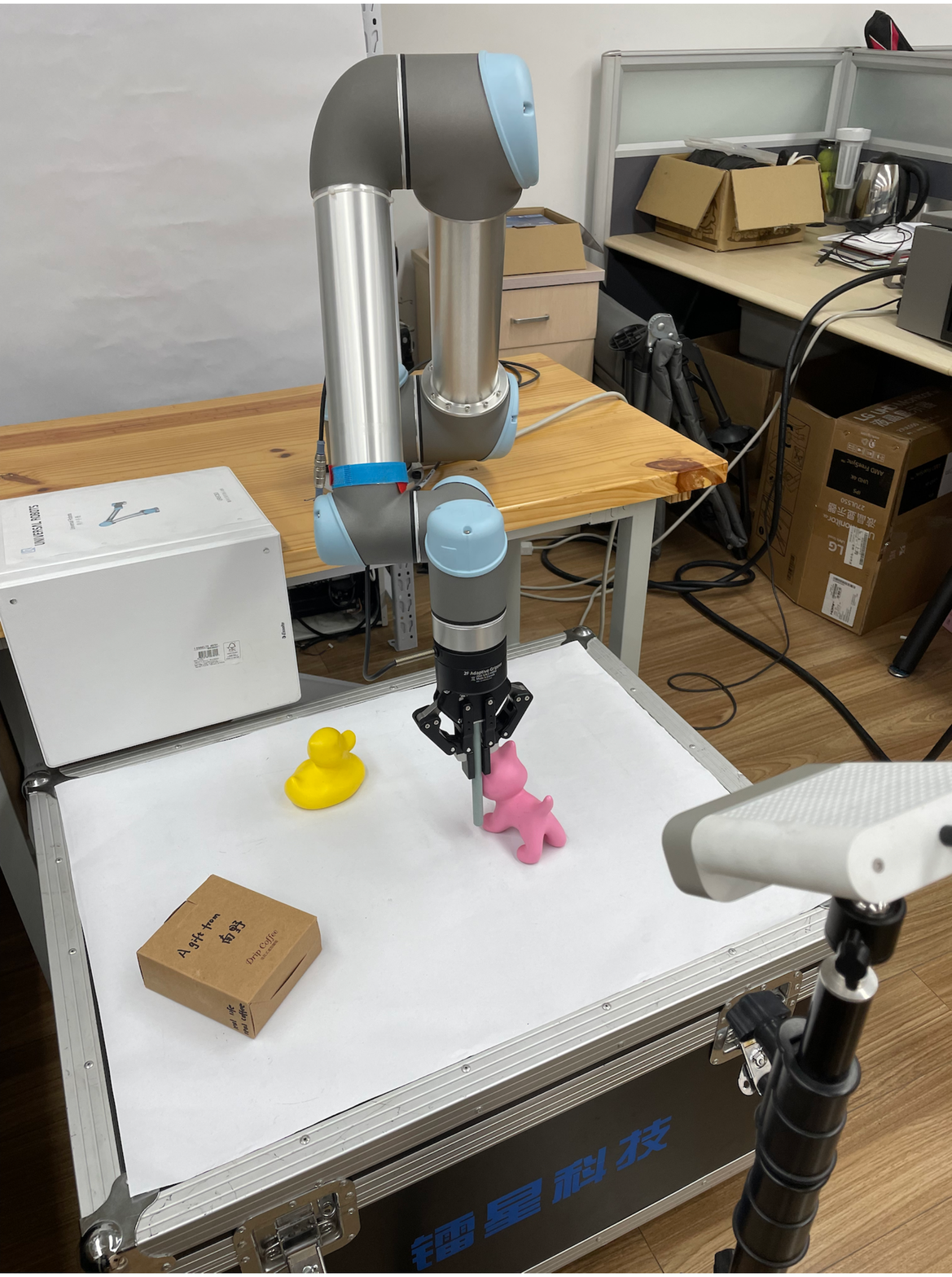}}
}
    \vspace{0.5em}
    \caption{\textbf{The real-world setup.} Poking is achieved via a UR5 robot and the RGB-D video is recorded by a Kinect camera.}
    \label{fig:env_setup}
\end{figure}
    The real-world setup is composed of a UR5 robot with a cylinder pusher tool and an RGB-D Kinect camera visualized in Fig.~\ref{fig:env_setup}.
    The relative pose between the camera and the base of the robot arm $T_{ca}$ is computed as follows:
    \begin{equation}
        T_{ca}=T_{gc}^{-1}T_{ga},
    \end{equation}
    where $T_{gc}$ is the relative pose between the gripper and the camera and $T_{ga}$ is the relative pose between the gripper and the base of the robot arm.
    In practice, $T_{ga}$ is read from the Robot Operating System (ROS).
    $T_{gc}$ is obtained by aligning the model of the gripper and an RGB-D image.
    The alignment is achieved via differentiable rendering.
    Representing the 3D model of the gripper as $G$ and an RGB-D image with the manually-annotated segmentation mask of the gripper as $M$, $T_{gc}$ is obtained by optimizing the following objective function:
    \begin{equation}
        \mathcal{L}(R,t)=\sum(\pi(RG+t)-M)^2,
    \end{equation}
    where $\pi$ is a differentiable neural renderer~\cite{kato2018neural} that produce the mask of the gripper.

    \section{Supplementary video}\label{sec:supplementary-video}
    The supplementary video provides a demonstration of the real-world robotic grasping task.
    It starts by describing the task, followed by the recorded video of the poking process, and ends with the application of object pose estimation and robotic grasping to demonstrate the practicality of our method in real-world scenarios.
    \bibliography{supp}